\pdfoutput=1

\documentclass[11pt]{article}

\usepackage[final]{acl}

\usepackage{times}
\usepackage{latexsym}

\usepackage[T1]{fontenc}
\usepackage[utf8]{inputenc}

\usepackage{microtype}
\usepackage{subcaption}
\usepackage{inconsolata}

\usepackage{graphicx}
\usepackage{booktabs}
\usepackage{multicol}
\usepackage{multirow}
\usepackage[many]{tcolorbox}

\usepackage{amsmath}
\usepackage{amssymb}
\usepackage{mathtools}
\usepackage{amsthm}
\usepackage{cleveref}
\theoremstyle{plain}
\newtheorem{theorem}{Theorem}[section]

\theoremstyle{definition}

\theoremstyle{remark}

\usepackage{amsmath,amsfonts,bm}

\def\eqref#1{equation~\ref{#1}}

\def\1{\bm{1}}

\DeclareMathAlphabet{\mathsfit}{\encodingdefault}{\sfdefault}{m}{sl}
\SetMathAlphabet{\mathsfit}{bold}{\encodingdefault}{\sfdefault}{bx}{n}

\def\gO{{\mathcal{O}}}

\newcommand{\vct}[1]{\pmb{#1}}

\newcommand{\alg}{\text{SNNE}}
\newcommand{\walg}{\text{WSNNE}}
\newcommand{\se}{\text{SE}}
\newcommand{\dse}{\text{DSE}}
\newcommand{\kleheat}{$\text{KLE}_{\text{heat}}$}
\newcommand{\klefull}{$\text{KLE}_{\text{full}}$}
\newcommand{\lexsim}{\text{LexSim}}
\newcommand{\sumeigv}{\text{SumEigv}}
\newcommand{\degree}{\text{Deg}}
\newcommand{\ecc}{\text{Eccen}}
\newcommand{\naive}{\text{NE}}

\newcommand{\ptrue}{\text{pTrue}}
\newcommand{\numset}{\text{NumSet}}
\newcommand{\name}{{Semantic Nearest Neighbor Entropy}}
\newcommand{\llamatwoseven}{\text{Llama2-7B}}
\newcommand{\llamatwothirdteen}{\text{Llama2-13B}}
\newcommand{\llamathree}{\text{Llama-3.1-8B}}
\newcommand{\phithree}{\text{Phi-3-mini}}
\newcommand{\gemmatwo}{\text{gemma-2-2b}}
\newcommand{\mistral}{\text{Mistral-Nemo}}
\newcommand{\squad}{\text{SQuAD}}
\newcommand{\triviaqa}{\text{TriviaQA}}
\newcommand{\nq}{\text{NaturalQuestion}}
\newcommand{\svamp}{\text{Svamp}}
\newcommand{\bioasq}{\text{BioASQ}}

\newcommand{\xsum}{\text{XSUM}}
\newcommand{\aeslc}{\text{AESLC}}
\newcommand{\wmtde}{\text{WMT-14 de-en}}
\newcommand{\wmtfr}{\text{WMT-14 fr-en}}
\newcommand{\auroc}{\text{AUROC}}
\newcommand{\auarc}{\text{AUARC}}
\newcommand{\prr}{\text{PRR}}
\newcommand{\rouge}{\text{ROUGE-L}}
\newcommand{\bert}{\text{BERTScore}}
\newcommand{\entail}{\text{entail}}
\newcommand{\embed}{\text{embed}}

\title{Beyond Semantic Entropy: Boosting LLM Uncertainty Quantification with Pairwise Semantic Similarity}

\author{Dang Nguyen \\
  UCLA CS \\
  \texttt{dangnth@cs.ucla.edu} \\\And
  Ali Payani \\
  Cisco Systems Inc. \\
  \texttt{apayani@cisco.com} \\\And
  Baharan Mirzasoleiman \\
  UCLA CS \\
  \texttt{baharan@cs.ucla.edu}
}

\begin{document}
\maketitle
\begin{abstract}
Hallucination in large language models (LLMs) can be detected by assessing the uncertainty of model outputs, typically measured using entropy. Semantic entropy (SE) enhances traditional entropy estimation by quantifying uncertainty at the semantic cluster level. However, as modern LLMs generate longer one-sentence responses, SE becomes less effective because it overlooks two crucial factors: intra-cluster similarity (the spread within a cluster) and inter-cluster similarity (the distance between clusters).
To address these limitations, we propose a simple black-box uncertainty quantification method inspired by nearest neighbor estimates of entropy. Our approach can also be easily extended to white-box settings by incorporating token probabilities. Additionally, we provide theoretical results showing that our method generalizes semantic entropy. Extensive empirical results demonstrate its effectiveness compared to semantic entropy across two recent LLMs (Phi3 and Llama3) and three common text generation tasks: question answering, text summarization, and machine translation. Our code is available at \href{https://github.com/BigML-CS-UCLA/SNNE}{https://github.com/BigML-CS-UCLA/SNNE}.
\end{abstract}

\section{Introduction}\label{sec:intro}
Large Language Models (LLMs) have demonstrated impressive capabilities in understanding and generating human-like text, revolutionizing various fields~\cite{bubeck2023sparks, team2024gemini}. However, they are not without flaws, and one of the most significant challenges is hallucination, i.e., incorrect or fabricated information that appears plausible~\cite{maynez-etal-2020-faithfulness,ji2023survey}. Detecting and mitigating hallucination is critical to ensuring the reliability and safety of LLMs.
A common approach to addressing this issue is uncertainty quantification (UQ)~\cite{huang2024survey}. By measuring the uncertainty of an LLM's outputs, we can identify potentially hallucinated content and flag it for further review, improving both accuracy and user trust.

A straightforward way to estimate uncertainty in LLMs is to generate multiple responses and aggregate their token-level likelihoods~\cite{malinin2020uncertainty}. 
However, this approach ignores semantic information, treating reworded yet equivalent answers as distinct.
Semantic Entropy (\se) \cite{farquhar2024detecting} addresses this by clustering semantically similar outputs using bidirectional entailment predictions from a NLI model before computing entropy. While effective for short responses, \se~struggles when model generates long one-sentence outputs, a pattern inherent to tasks such as summarization~\cite{zhang2020pegasus} and translation~\cite{hendy2023good}.
In these cases, \se~often reverts to the previous naive approach while its discrete version, Discrete Semantic Entropy (\dse), yields constant values. These shortcomings highlight the need for more robust uncertainty estimation techniques tailored to complexities of lengthy model responses.

\begin{figure*}[t!]
    \centering
    \begin{subfigure}[b]{0.45\textwidth}
        \includegraphics[width=0.8\columnwidth]{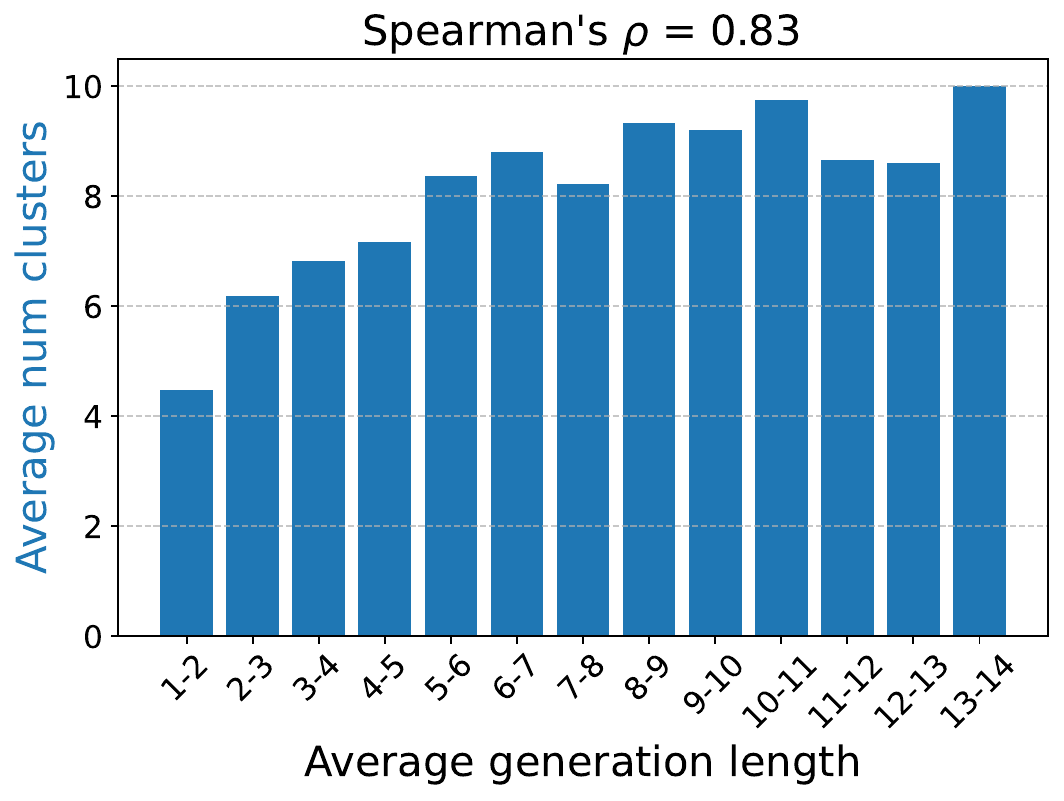}
    \end{subfigure}
    \begin{subfigure}[b]{0.45\textwidth}
        \includegraphics[width=0.95\columnwidth]{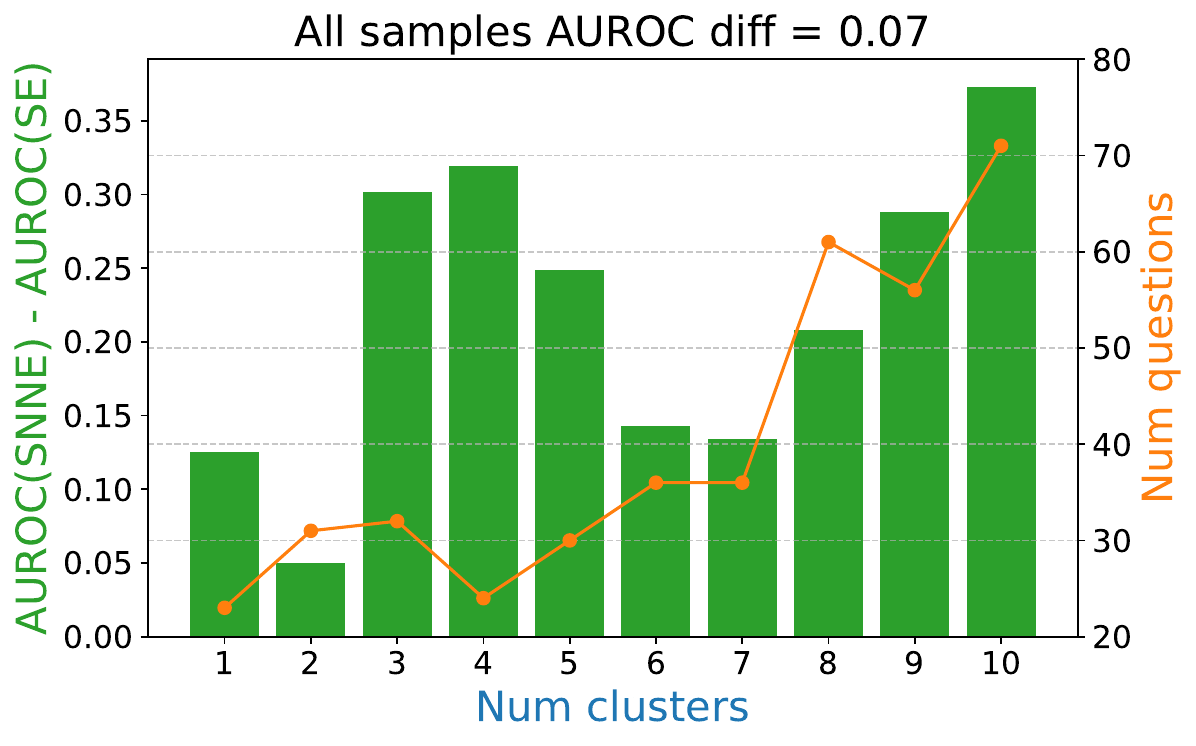}
    \end{subfigure}
    \vspace{-2mm}
    \caption{\textbf{Generated answers of \llamathree~on \squad.}
    (left) The Spearman's rank correlation coefficient between the average output length and the number of semantic clusters is 0.83, indicating a high correlation. 
    (right) The AUROC difference between \alg~and \se~when calculating on subsets of questions with different number of clusters. \alg~is consistently more distinctive than \se, especially when the number of clusters is large.
    }
    \label{fig:squad_llama3}
    \vspace{-4mm}
\end{figure*}

We empirically observe that state-of-the-art LLMs tend to generate longer one-sentence outputs on common QA benchmarks: 
\llamathree~and \phithree~produce responses with an average length of 4.1 and 4.9 words, respectively, compared to 2.3 words for \llamatwoseven.
In these scenarios, model outputs are more likely to belong to different semantic groups, thus capturing inter-cluster similarity is important. In addition, each semantic cluster has different spread, necessitating the consideration of inter-cluster similarity. To address these two issues, we propose a simple yet effective black-box UQ method inspired by nearest neighbor estimates of entropy~\cite{beirlant1997nonparametric}. Our approach can be seamlessly extended to white-box settings by incorporating token probabilities, providing flexibility across different use cases. Additionally, we theoretically prove that our method generalizes semantic entropy, offering greater expressiveness. Extensive experiments on nine datasets and state-of-the-art LLMs demonstrate that our method consistently outperforms \se~and other baselines across question answering, summarization, and translation.

\section{Existing uncertainty quantification methods}\label{sec:related_works}

\textbf{Black-box methods.} 
These methods rely solely on generated answers. \ptrue~\cite{kadavath2022language} estimates uncertainty by prompting the model a True/False question to verify its own responses. The model’s uncertainty is then determined by the probability of selecting False. Lexical Similarity (\lexsim) \cite{fomicheva2020unsupervised} measures uncertainty via average lexical overlap among responses but struggles with semantically similar outputs that use different wording. Graph-based approaches~\cite{lin2024generating} structure responses as a similarity graph and derive uncertainty from its properties. \numset~counts semantic clusters found by \se, while \sumeigv~generalizes this via the sum of eigenvalues of the graph Laplacian. \degree~computes the average pairwise semantic similarities, and \ecc~quantifies uncertainty based on distances from responses to their center in the space defined by the $k$ smallest eigenvectors of the graph Laplacian. Because graph-based methods use summation to aggregate  answer similarities, they are sensitive to outliers or peculiar answers. LUQ~\cite{zhang2024luq} is designed for multi-sentence scenarios by aggregating atomic uncertainty scores computed for individual sentences. However, its atomic scoring relies on NLI-based similarity, which inherits the limitations of LexSim and graph-based methods. Our method can be integrated into LUQ to provide more reliable atomic scores, offering a promising direction for extending our approach to multi-sentence settings. We leave this for future work.
\looseness=-1

\textbf{White-box methods.} 
These methods utilize internal model information, such as token likelihoods and representations, to assess uncertainty. Naive Entropy (\naive)~\cite{malinin2020uncertainty} computes entropy over length-normalized sequence probabilities but is sensitive to variations in probability assignments for semantic equivalence answers, leading to unreliable estimates.
Semantic Entropy (\se)~\cite{farquhar2024detecting} improves robustness by leveraging a greedy bi-directional entailment clustering algorithm to group responses into semantic clusters before computing entropy but does not account for semantic distances between generations.
Kernel Language Entropy (KLE)~\cite{nikitin2024kernel} addresses this by using von Neumann entropy on a semantic kernel. 
However, the use of heat and Matern kernels for constructing semantic kernels in KLE makes it challenging to interpret the model’s uncertainty and may lead to the loss of information. Empirically, we show that directly utilizing semantic similarities between answers leads to improved performance and more accurate uncertainty estimation. 
In addition, its $\gO(N^3)$ complexity makes it impractical for large-scale uncertainty estimation. In contrast, SE provides a more computationally efficient approach $\gO(N^2)$ but struggles with long outputs. Given the efficiency and intuitive formulation, SE serves as a foundation for further improvement, motivating our approach to address its limitations.\looseness=-1

Rather than relying on clustering, SAR~\cite{duan2023shifting} computes a soft aggregation of word- or sentence-level probabilities weighted by their semantic similarity. It also reduces the impact of irrelevant tokens and low-quality sequence samples. Although SAR achieved strong performance on the LM-Polygraph benchmark~\cite{vashurin2025benchmarking}, its effectiveness degrades with increasing output length, as shown in their Table 4.
EigenScore~\cite{chen2024inside}, in contrast, assumes deeper access to the model’s internal states. It leverages the eigenvalues of the covariance matrix of output embeddings—a more computationally intensive approach compared to methods based on logits or probabilities. Additionally, EigenScore applies feature clipping to suppress extreme values in the embedding space, aiming to avoid overconfident, self-consistent hallucinations. However, this technique requires tuning a clipping threshold, which can introduce additional complexity.

\section{Preliminaries}

\textbf{Notations.} Let $P(\vct{a} | \vct{q})$ denote the sequence probability that the model generates the answer $\vct{a}$ given question $\vct{q}$, i.e, $P(\vct{a} | \vct{q}) = \sum_j \log p(\vct{a}_j | \vct{q} \oplus \vct{a}_{<j})$. The length-normalized sequence probability is defined as 
$\tilde{P}(\vct{a} | \vct{q}) = P(\vct{a} | \vct{q}) / len(\vct{a})$.

\textbf{Uncertainty quantification pipeline.} In general, each UQ method consists of two main steps.

1. Generation: Given question $\vct{q}$, sample $n$ answers $\vct{a}^1, \ldots, \vct{a}^n$ from LLM.

2. Entropy estimation: Compute uncertainty based on question $\vct{q}$ and answers $\{ \vct{a}^i\}_{i=1}^n$.

\textbf{Semantic Entropy.} \se~\cite{farquhar2024detecting} uses a bidirectional entailment model to cluster outputs into $M$ semantic classes $\{ C_k\}_{k=1}^M$. Let the semantic class probability be the sum of sequence probabilities of all outputs in that class, i.e., $P(C_k) = \sum_{i, \vct{a}^i \in C_k} \tilde{P}(\vct{a}^i | \vct{q})$. Then, define the normalized semantic class probability as $\bar{P}(C_k) = \frac{P(C_k)}{\sum_{j=1}^M P(C_j)}$. Using the Rao-Blackwellized Monte Carlo estimator, SE computes the entropy as
\begin{align}
    \se(\vct{q}) &= - \sum_{k=1}^M \bar{P}(C_k) \log \bar{P}(C_k) \label{eq:semantic_entropy}
\end{align}

\textbf{Discrete Semantic Entropy.} \dse~\cite{farquhar2024detecting} is an extension of \se~to the black-box setting by approximating the normalized semantic class probability with the empirical cluster probability, i.e., 
$\frac{|C_k|}{n}$.
\vspace{-3mm}
Then, the formula of \dse~reads
\begin{align}
    \dse(\vct{q}) &= - \sum_{i=1}^M \frac{|C_k|}{n} \log \frac{|C_k|}{n}
    \label{eq:discrete_semantic_entropy}
\end{align}
\vspace{-3mm}

\textbf{Issues of Semantic Entropy.} Empirically, we observe that state-of-the-art LLMs tend to generate longer responses than their predecessors. For instance,  on five QA datasets in our experiments, \llamathree~and \phithree~produce responses with an average length of 4.1 and 4.9 words, respectively, compared to 2.3 words for \llamatwoseven. Additionally, as shown in Figure~\ref{fig:squad_llama3} left, the number of semantic clusters is strongly correlated with response length.
When the number of semantic clusters $M$ approaches $n$, \dse~produces a constant entropy regardless of $\{ \vct{a}^i\}_{i=1}^n$ because it fails to account for the similarity among clusters (inter-cluster similarity). Even if the number of clusters $M$ is small in Figure~\ref{fig:squad_llama3} right, \se~falls behind our method as it lacks the spread of different clusters (intra-cluster similarity) in its formulation. These two issues make (D)\se~render indistinctive entropy to detect hallucination.
\section{Incorporating intra-and inter-cluster similarity in uncertainty quantification}\label{sec:method}

We can mitigate the information missing issues of (D)\se~by leveraging both the intra-and inter-cluster similarities between generated answers, leading to more accurate uncertainty estimation. To alleviate the effect of outliers, we leverage LogSumExp operation to aggregate similarity. Putting together, \textit{without the need of clustering}, we define \name~(\alg) as
\vspace{-3mm}
\begin{equation}\label{eq:alg}
\alg(\vct{q}) = - \frac{1}{n} \sum_{i=1}^n \log \sum_{j=1}^n \exp\left(\frac{f(\vct{a}^i, \vct{a}^j | \vct{q})}{\tau}\right)
\end{equation}
where $f$ measures the similarity between two answers given the question. 
The inner summation in Eq~\ref{eq:alg} effectively accounts for both intra- and inter-cluster similarities without requiring clustering, as SE does. Instead of explicitly grouping outputs into clusters, $f$ naturally captures intra-cluster similarity when $\vct{a}^i$ and $\vct{a}^j$ belong to the same semantic group and inter-cluster similarity when they do not.
Because LogSumExp operation is a smooth approximation to the maximum function, \alg~resembles the entropy estimation based on the nearest neighbor distances~\cite{beirlant1997nonparametric}. Thus, our method is less sensitive to outliers compared to \lexsim~and graph-based approaches.

\textbf{Extending \alg~to white-box settings.} 
Token probabilities of the generated outputs can be incorporated to weight the summation in Eq~\ref{eq:alg}, enhancing the method’s sensitivity to model confidence. 
Based on this, we propose the white-box version of \alg~as:
\vspace{-1mm}
\begin{align}\label{eq:walg}
    &\walg(\vct{q}) \nonumber \\
    &= - \sum_{i=1}^n \bar{P}(\vct{a}^i | \vct{q}) 
    \log \sum_{j=1}^n \exp\left(\frac{f(\vct{a}^i, \vct{a}^j | \vct{q})}{\tau}\right)
\end{align}
where $\bar{P}(\vct{a}^i | \vct{q}) = \tilde{P}(\vct{a}^i | \vct{q}) / {\sum_{j=1}^n \tilde{P}(\vct{a}^j | \vct{q})}$

\begin{figure*}[t!]
    \centering
    \includegraphics[width=0.85\textwidth]{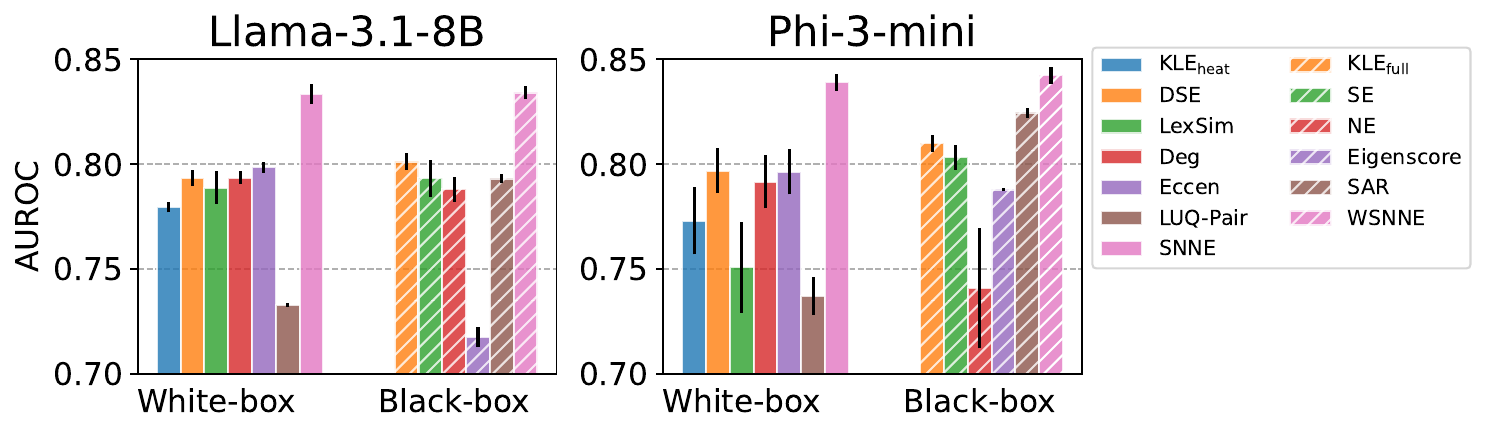}
    \vspace{-2mm}
    \caption{Average AUROC score of \llamathree~and \phithree~on 5 QA tasks.}
    \label{fig:main_qa_auroc}
    \vspace{-3mm}
\end{figure*} 

\begin{figure*}[t!]
    \centering
    \includegraphics[width=0.85\textwidth]{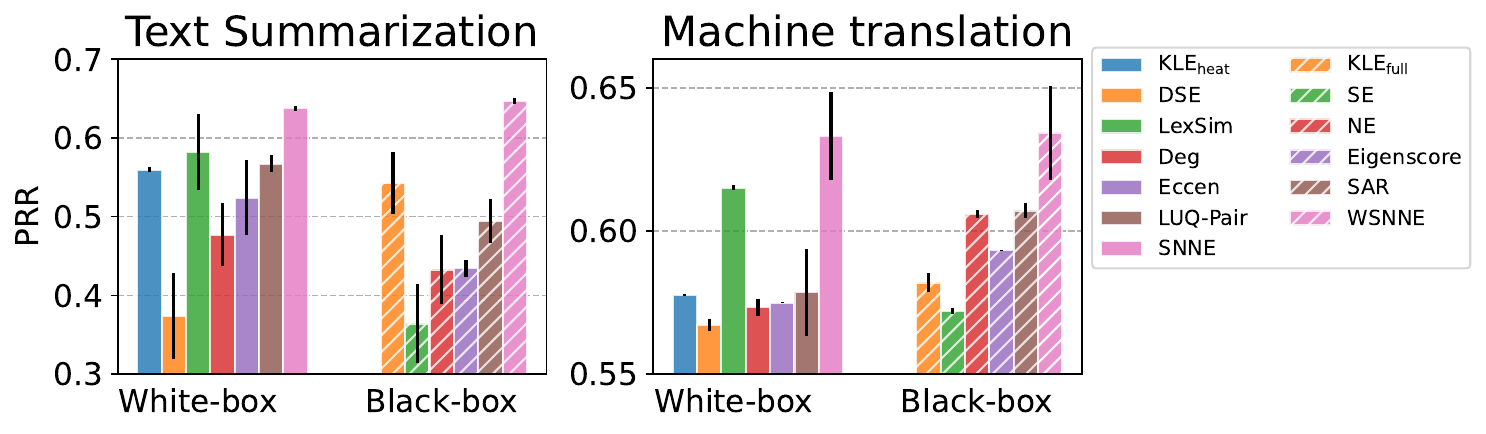}
    \vspace{-2mm}
    \caption{Average PRR score of \phithree~on 2 text summarization and 2 machine translation tasks.}
    \label{fig:main_ts_mt_prr}
    \vspace{-4mm}
\end{figure*} 

\begin{figure}[t!]
    \centering
    \includegraphics[width=0.9\linewidth]{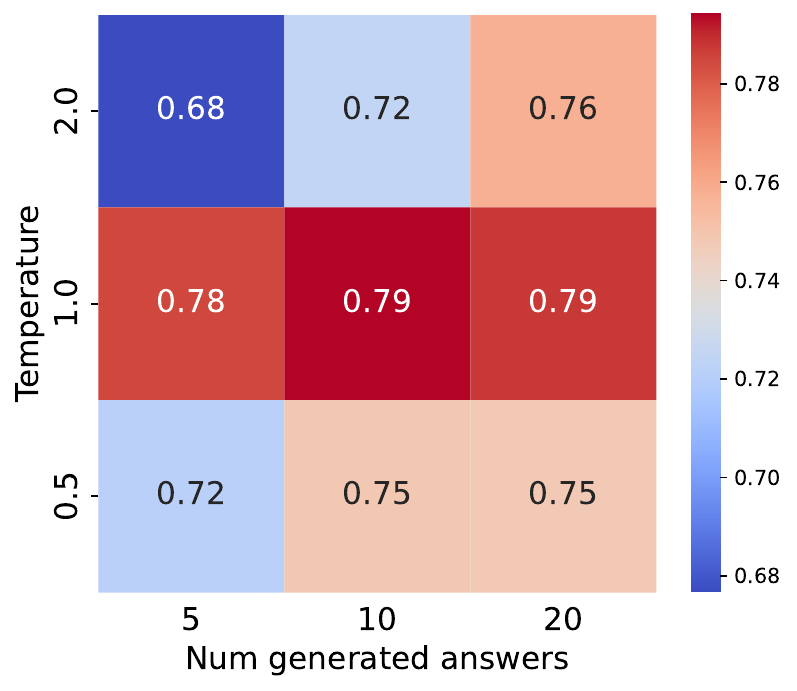}
    \caption{Effect of number of generated answers and generation temperature on the performance of \alg. We measure the \auroc~score of \llamathree~on \squad~dataset.}
    \label{fig:change_num_gen_temp}
    \vspace{-3mm}
\end{figure} 

\textbf{Choices of similarity function.} The similarity function $f$ can be selected from several options. The first choice is \rouge~\cite{lin-2004-rouge} used in \lexsim~\cite{fomicheva2020unsupervised}. Another option is to use the predicted scores from NLI models, as in graph-based methods~\cite{lin2024generating}. Finally, $f$ can also be defined as the cosine similarity between sentence embeddings of the model outputs.\looseness=-1

\textbf{Theoretical results.}
The following theorem shows that if we neglect inter-cluster similarity by assigning zero similarity to answers from different sematic clusters and intra-cluster similarity by assigning a constant value to answers within the same clusters, \alg~recovers \dse.

\begin{theorem}\label{thm:dse}
Let function $f$ be defined as
$ f(\vct{a}^i, \vct{a}^j | \vct{q}) = \tau \log(\frac{1}{n})$ if $\vct{a}^i, \vct{a}^j \in C_k $, otherwise $-\infty$.
\alg~is equivalent to \dse~as defined in Eq~\ref{eq:discrete_semantic_entropy}.
\end{theorem}

Instead of a constant value, if we assign a term based solely on the sequence probability, \walg~recovers \se.
\begin{theorem}\label{thm:se}
Let function $f$ be defined as
$f(\vct{a}^i, \vct{a}^j | \vct{q}) = \tau \log(\frac{\tilde{P}(\vct{a}^j | \vct{q})}{Q})$ if $\vct{a}^i, \vct{a}^j \in C_k$, otherwise $-\infty$
where $Q = \sum_{i=1}^n \tilde{P}(\vct{a}^i | \vct{q})$. 
\walg~is equivalent to \se~as defined in Eq ~\ref{eq:semantic_entropy}.
\end{theorem}

The proofs of Theorems~\ref{thm:dse} and~\ref{thm:se} are provided in Appendix~\ref{app:proofs}. As stated, (D)SE is a special case of (W)SNNE under specific similarity metrics. Specifically, Theorem~\ref{thm:dse} shows that when a constant similarity is assigned to examples within a cluster and inter-cluster interactions are ignored, SNNE simplifies to DSE. Similarly, Theorem~\ref{thm:se} demonstrates that when intra-cluster interactions are disregarded, WSNNE reduces to SE. These results imply that (W)SNNE provides a more expressive measure of uncertainty than (D)SE, and that leveraging informative similarity metrics enables (W)SNNE to outperform (D)SE, as evidenced by our experiments. A similar form of generalization was also observed in KLE’s framework~\cite{nikitin2024kernel}.

Furthermore, our method can asymptotically ($\tau \to 0$) recover LUQ-Pair when $f(\vct{a}^i, \vct{a}^j | \vct{q}) = \tau P(\text{entail} | \vct{a}^i, \vct{a}^j)$. Therefore, our method can be integrated into their method to find better atomic scores. Empirically, we showed that SNNE outperforms LUQ-PAIR for one-sentence generation, highlighting the better atomic uncertainty estimation of our method.
\section{Experiments}\label{sec:experiments}

\subsection{Settings}
\textbf{Models.} We use Llama-3.1-8B~\citep{dubey2024llama} and 
Phi-3-mini-4k-instruct~\citep{abdin2024phi}.

\textbf{Datasets.} We evaluate our method across three NLP tasks: question answering (QA), text summarization (TS), and machine translation (MT). For QA, we use five datasets following SE~\cite{farquhar2024detecting} while for TS and MT, we adopt two summarization and two translation datasets in LM-Polygraph~\citep{fadeeva-etal-2023-lm}. Details are deferred to Appendix~\ref{app:exp_details}.

\textbf{Evaluation metrics.} For QA tasks, we assess UQ methods using \auroc~and \auarc~\cite{nadeem2009accuracy}, following~\citet{lin2024generating}. For TS and MT tasks, we adopt \prr~\cite{malinin-etal-2017-incorporating} in line with LM-Polygraph. 

\textbf{Baselines.} We compare our methods with white-box and black-box UQ methods in Section~\ref{sec:related_works}.

Additional details are given in Appendix~\ref{app:exp_details}.

\subsection{Results}
\textbf{QA datasets.} Figure~\ref{fig:main_qa_auroc} shows the performance of various uncertainty quantification (UQ) methods on answers generated by \llamathree~and \phithree. Both \alg~and \walg~consistently outperform existing white-box and black-box baselines by a significant margin across both models. Among white-box methods, \klefull~ranks second on \llamathree, yet its black-box counterpart, \kleheat, performs worse than the simpler \dse. Notably, our approach surpasses SAR, the previous state-of-the-art on the LM-Polygraph benchmark~\cite{vashurin2025benchmarking}, especially in cases involving longer, single-sentence outputs where SAR’s performance declines.

\textbf{TS and MT datasets.} Figure~\ref{fig:main_ts_mt_prr} demonstrates the superiority of our methods in terms of PRR score. \lexsim~is the runner-up in the black-box setting though its QA performance is much worse than the other baselines. For very long generations, the number of overlapping words in \lexsim~is a good indicator of semantic similarity as we confirmed in Table~\ref{tab:change_sim} in Appendix~\ref{app:exp_results}.  Notably, our methods improve over (D)\se~and KLE by a clear margin. This highlights the importance of incorporating intra-and inter-cluster similarities explicitly into entropy estimation in the long-generation cases.

\textbf{Effect of number of generations and generation temperature.} Figure~\ref{fig:change_num_gen_temp} shows the impact of the number of model outputs and temperature on the \auroc~score of \alg. Increasing the number of generations enhances performance but comes at a higher inference cost. Additionally, at low temperatures, performance saturates at around 10 outputs. Using either too low (0.5) or too high (2.0) temperature degrades performance, as overly conservative or excessively diverse outputs compromise the quality of entropy estimation. Note that our generation setting (temperature 1.0 and 10 generated answers) is directly adopted from SE's paper~\cite{farquhar2024detecting}.

\textbf{Additional results.} Results for other models and ablation studies on the similarity function, scale factor $\tau$ can be found in Appendix~\ref{app:exp_results}.
\section{Conclusion}\label{sec:conclusion}

In this paper, we introduced a novel black-box uncertainty quantification method to address the limitations of semantic entropy in detecting hallucinations in long-generation scenarios. Our approach effectively accounts for both intra-and inter-cluster similarities and mitigate outliers, which are critical for accurate uncertainty estimation in modern LLMs. We also demonstrated that our method can be extended to white-box settings and provided theoretical results showing its generalization of semantic entropy. Extensive experiments on multiple LLMs and text generation tasks show that our method consistently outperforms existing uncertainty quantification methods.
\section{Limitations}\label{sec:limit}
In this paper, we did not investigate uncertainty estimation in cases where the model generates multiple sentences or an entire paragraph. A naive approach would be to compute the uncertainty for each sentence independently and then aggregate these values into a single scalar. We leave this direction for our future work. Additionally, for different data formats such as mathematical expressions, LaTeX equations, or code, our method requires further considerations. Designing an appropriate similarity function could help generalize our approach to these types of data. Finally, our method, similar to other existing UQ methods, requires sample multiple answers to estimate entropy, incurring additional inference cost.

\section*{Acknowledgements}
This research was partially supported by the National Science Foundation CAREER Award 2146492 and Cisco Systems.

\bibliography{custom}

\newpage
\appendix
\begin{table*}[t!]
    \centering
    \caption{Instruction prompts for different tasks.}
    \label{tab:instruct_prompt}
    \begin{tabular}{c|c}
        \textbf{Task} & \textbf{Instruction} \\
        \hline
        QA & Answer the following question as briefly as possible. \\
        \hline
        \xsum & Here's the text and it's short one-sentence summary. \\
        \hline
        \aeslc & Write a short subject line for the email. Output only the subject line itself. \\
        \hline
        \wmtde & Here is a sentence in German language and its translation in English language. \\
        \hline
        \wmtfr & Here is a sentence in French language and its translation in English language. \\
    \end{tabular}
\end{table*}

\begin{table*}[!t]
    \caption{Average \auroc~score on 5 QA tasks.}
    \label{tab:qa_auroc_no_math}
    \centering
    \scalebox{0.7}{
        \begin{tabular}{c|c|cccc|cccccccccc}
        \toprule
         \multirow{2}{*}{Model} & \multirow{2}{*}{Acc} & \multicolumn{4}{|c|}{White-box} & \multicolumn{9}{|c}{Black-box} \\
         \cline{3-15}
         && \klefull & \se & \naive & \walg & \kleheat & \dse & \ptrue & \numset & \lexsim & \sumeigv & \degree & \ecc & \alg \\
         \hline\hline
         \llamatwoseven & 41.73 & {0.80} & 0.79 & 0.73 & \textbf{0.81} & 0.78 & 0.79 & 0.67 & 0.79 & 0.79 & 0.79 & 0.79 & 0.79 & \textbf{0.80} \\
         \hline
         \llamatwothirdteen & 46.07 & 0.79 & 0.78 & 0.74 & \textbf{0.80} & 0.78 & 0.78 & 0.74 & 0.78 & 0.78 & 0.78 & 0.78 & 0.78 & \textbf{0.80} \\
         \hline
         \llamathree & 50.76 & 0.80 & 0.79 & 0.79 & \textbf{0.83} & 0.78 & 0.79 & 0.72 & 0.79 & 0.79 & 0.79 & 0.79 & 0.80 & \textbf{0.83} \\
         \hline
         \phithree & 42.84 & 0.81 & 0.80 & 0.74 & \textbf{0.84} & 0.77 & 0.80 & 0.56 & 0.79 & 0.75 & 0.77 & 0.79 & 0.80 & \textbf{0.84} \\
         \hline
         \gemmatwo & 39.28 & 0.83 & 0.82 & 0.77 & \textbf{0.84} & 0.82 & 0.82 & 0.72 & 0.82 & 0.80 & 0.81 & 0.82 & 0.82 & \textbf{0.84} \\
         \hline
         \mistral & 54.02 & 0.80 & 0.79 & 0.75 & \textbf{0.82} & 0.78 & 0.79 & 0.74 & 0.79 & 0.78 & 0.78 & 0.79 & 0.79 & \textbf{0.81} \\
         \bottomrule
        \end{tabular}
    }
\end{table*}

\begin{table*}[!t]
    \caption{Average \auarc~score on 5 QA tasks.}
    \label{tab:qa_auarc_no_math}
    \centering
    \scalebox{0.7}{
        \begin{tabular}{c|c|cccc|cccccccccc}
        \toprule
         \multirow{2}{*}{Model} & \multirow{2}{*}{Acc} & \multicolumn{4}{|c|}{White-box} & \multicolumn{9}{|c}{Black-box} \\
         \cline{3-15}
         && \klefull & \se & \naive & \walg & \kleheat & \dse & \ptrue & \numset & \lexsim & \sumeigv & \degree & \ecc & \alg \\
         \hline\hline
         \llamatwoseven & 41.73 & \textbf{0.63} & 0.61 & 0.59 & \textbf{0.63} & 0.62 & 0.61 & 0.54 & 0.61 & 0.61 & 0.62 & 0.61 & 0.62 & \textbf{0.63} \\
         \hline
         \llamatwothirdteen & 46.07 & \textbf{0.67} & 0.65 & 0.63 & 0.66 & 0.65 & 0.64 & 0.63 & 0.63 & 0.66 & 0.64 & 0.63 & 0.65 & \textbf{0.67} \\
         \hline
         \llamathree & 50.76 & 0.73 & 0.71 & 0.72 & \textbf{0.74} & 0.71 & 0.71 & 0.66 & 0.70 & 0.71 & 0.71 & 0.71 & 0.71 & \textbf{0.74} \\
         \hline
         \phithree & 42.84 & {0.64} & 0.63 & 0.60 & \textbf{0.66} & 0.63 & 0.63 & 0.45 & 0.63 & 0.61 & 0.62 & 0.63 & 0.63 & \textbf{0.66} \\
         \hline
         \gemmatwo & 39.28 & \textbf{0.64} & 0.62 & 0.59 & {0.63} & 0.62 & \textbf{0.63} & 0.54 & 0.62 & 0.60 & 0.61 & 0.62 & 0.62 & \textbf{0.63} \\
         \hline
         \mistral & 54.02 & \textbf{0.75} & 0.72 & 0.70 & {0.74} & 0.72 & 0.72 & 0.70 & 0.72 & 0.72 & 0.72 & 0.72 & \textbf{0.73} & \textbf{0.73} \\
         \bottomrule
        \end{tabular}
    }
\end{table*}

\begin{table*}[!t]
    \caption{Average \prr~score on 2 summarization tasks (\xsum, \aeslc) and 2 translation tasks (\wmtde, \wmtfr). For correctness metric, R denotes \rouge~and B denotes \bert.}
    \label{tab:ts_mt_prr}
    \centering
    \scalebox{0.65}{
        \begin{tabular}{c|c|c|cccc|ccccccccc}
        \toprule
         \multirow{2}{*}{Model} & \multirow{2}{*}{Metric} & \multirow{2}{*}{Score} & \multicolumn{4}{|c|}{White-box} & \multicolumn{9}{|c}{Black-box} \\
         \cline{4-16}
         &&& \klefull & \se & \naive & \walg & \kleheat & \dse & \ptrue & \numset & \lexsim & \sumeigv & \degree & \ecc & \alg \\
         \hline\hline
         \multicolumn{16}{c}{Summarization} \\
         \hline\hline
         \multirow{2}{*}{\phithree} & R & 0.11 & 0.20 & 0.14 & 0.17 & \textbf{0.27} & 0.21 & 0.14 & 0.24 & 0.17 & {0.23} & 0.19 & 0.19 & 0.20 & \textbf{0.26} \\
         & B & 0.44 & 0.54 & 0.36 & 0.43 & \textbf{0.65} & 0.56 & 0.37 & 0.52 & 0.38 & {0.58} & 0.47 & 0.48 & 0.52 & \textbf{0.64} \\
         \hline\hline
         \multicolumn{16}{c}{Translation} \\
         \hline\hline
         \multirow{2}{*}{\phithree} & R & 0.62 & 0.58 & 0.57 & 0.60 & \textbf{0.63} & 0.58 & 0.57 & 0.56 & 0.56 & 0.61 & 0.57 & 0.57 & 0.58 & \textbf{0.63} \\
         & B & 0.92 & 0.71 & 0.70 & 0.73 & \textbf{0.75} & {0.71} & 0.70 & 0.70 & 0.70 & {0.73} & 0.70 & 0.71 & 0.71 & \textbf{0.75} \\
         \bottomrule
        \end{tabular}
    }
\end{table*}

\begin{table*}[!t]
    \caption{Effect of similarity function of \alg~2 summarization tasks (\xsum, \aeslc) and 2 translation tasks (\wmtde, \wmtfr).}
    \label{tab:change_sim}
    \centering
    \scalebox{0.75}{
        \begin{tabular}{c|c|ccc|ccc}
        \toprule
         \multirow{2}{*}{Model} & \multirow{2}{*}{Metric} & \multicolumn{3}{|c|}{Translation} & \multicolumn{3}{|c}{Summarization} \\
         \cline{3-8}
         && \embed & \entail & \rouge & \embed & \entail & \rouge \\
         \hline\hline
         \multirow{2}{*}{\phithree} & \rouge & 0.62 $\pm$ 0.02 & 0.60 $\pm$ 0.02 & \textbf{0.63} $\pm$ 0.02 & 0.22 $\pm$ 0.01 & 0.22 $\pm$ 0.01 & \textbf{0.26} $\pm$ 0.02 \\
         & \bert & \textbf{0.75} $\pm$ 0.01 & 0.73 $\pm$ 0.01 & \textbf{0.75} $\pm$ 0.01 & 0.56 $\pm$ 0.02 & 0.57 $\pm$ 0.01 & \textbf{0.64} $\pm$ 0.01 \\
         \bottomrule
        \end{tabular}
    }
\end{table*}

\begin{table*}[!t]
    \centering
    \begin{tabular}{c|c|c|c|c|c}
        \toprule
        Model & Task & Metric & $\tau$ & WSNNE & SNEE \\
        \midrule\midrule
        \multirow{4}{*}{Llama-3.1-8B} & \multirow{4}{*}{QA} & \multirow{4}{*}{AUROC} & 0.1 & 0.830 $\pm$ 0.003 & 0.830 $\pm$ 0.004 \\
        &&& 1 & \textbf{0.833 $\pm$ 0.002} & \textbf{0.832 $\pm$ 0.004} \\
        &&& 10 & 0.832 $\pm$ 0.003 & 0.830 $\pm$ 0.004 \\
        &&& 100 & 0.831 $\pm$ 0.003 & 0.830 $\pm$ 0.004 \\
        \midrule
        \multirow{4}{*}{Phi-3-mini} & \multirow{4}{*}{QA} & \multirow{4}{*}{AUROC} & 0.1 & 0.836 $\pm$ 0.003 & 0.832 $\pm$ 0.003 \\
        &&& 1 & \textbf{0.841 $\pm$ 0.004} & \textbf{0.838 $\pm$ 0.004} \\
        &&& 10 & 0.840 $\pm$ 0.004 & 0.836 $\pm$ 0.004 \\
        &&& 100 & 0.840 $\pm$ 0.004 & 0.834 $\pm$ 0.004 \\
        \midrule
        \multirow{4}{*}{Phi-3-mini} & \multirow{4}{*}{TS} & \multirow{4}{*}{PRR} & 0.1 & 0.645 $\pm$ 0.002 & 0.634 $\pm$ 0.002 \\
        &&& 1 & \textbf{0.646 $\pm$ 0.005} & \textbf{0.636 $\pm$ 0.005} \\
        &&& 10 & 0.645 $\pm$ 0.005 & 0.635 $\pm$ 0.005 \\
        &&& 100 & 0.645 $\pm$ 0.005 & 0.635 $\pm$ 0.005 \\
        \midrule
        \multirow{4}{*}{Phi-3-mini} & \multirow{4}{*}{MT} & \multirow{4}{*}{PRR} & 0.1 & 0.632 $\pm$ 0.015 & 0.631 $\pm$ 0.015 \\
        &&& 1 & \textbf{0.635 $\pm$ 0.016} & \textbf{0.632 $\pm$ 0.016} \\
        &&& 10 & 0.634 $\pm$ 0.016 & 0.632 $\pm$ 0.016 \\
        &&& 100 & 0.634 $\pm$ 0.016 & 0.632 $\pm$ 0.016 \\
        \midrule
        \bottomrule
    \end{tabular}
    \caption{Effect of the scale factor $\tau$ on 2 summarization tasks ( XSUM , AESLC ) and 2 translation tasks (WMT-14 de-en, WMT-14 fr-en).}
    \label{tab:scale_factor}
\end{table*}

\section{Proofs}\label{app:proofs}
\subsection{Proof of Theorem~\ref{thm:dse}}
\begin{proof}
    We define the similarity function $f$ as
    \begin{align}
        f(\vct{a}^i, \vct{a}^j | \vct{q}) = 
            \begin{cases} 
                \tau \log(\frac{1}{n}), & \text{if } \exists k \text{ s.t. } \vct{a}^i, \vct{a}^j \in C_k \\
                -\infty, & \text{otherwise}
            \end{cases} \label{eq:f_alg}
    \end{align}
    In other words, we give a constant similarity for answers belonging to the same semantic group and zero similarity for answers in different groups. Then, the total similarity between $\vct{a}^i$ to all answers $\{\vct{a}^j\}_{j=1}^n$ becomes $\frac{|C_k|}{n}$ where $C_k$ is the semantic group of $\vct{a}^i$. Therefore, Eq~\ref{eq:alg} reads
    \begin{align}
        \alg(\vct{q}) &= - \frac{1}{n} \sum_{i=1, \vct{a}^i \in C_k}^n \log \frac{|C_k|}{n} \\
        &= - \sum_{k=1}^M \frac{|C_k|}{n} \log \frac{|C_k|}{n} \\
        &= \dse(\vct{q})
    \end{align}
    The penultimate equality holds because $n = \sum_{k=1}^M |C_k|$.
\end{proof}

\subsection{Proof of Theorem~\ref{thm:se}}
\begin{proof}
    For a given question $\vct{q}$, let $Q$ be the total length-normalized sequence probability of its generated answers. We have the following equalities.
    \begin{align}
        Q &= \sum_{i=1}^n \tilde{P}(\vct{a}^i | \vct{q}) \\
        &= \sum_{k=1}^M P(C_k).
    \end{align}
    We define the similarity function $f$ as
    \begin{align}
        f(\vct{a}^i, \vct{a}^j | \vct{q}) = 
            \begin{cases} 
                \tau \log(\frac{\tilde{P}(\vct{a}^j | \vct{q})}{Q}), & \text{if } \vct{a}^i, \vct{a}^j \in C_k \\
                -\infty, & \text{otherwise}
            \end{cases} \label{eq:f_walg}
    \end{align}
    Similar to Eq~\ref{eq:f_alg}, we give a zero similarity for answers in different groups but give an asymmetric similarity based on the length-normalized sequence probability for answers belonging to the same semantic group. Then, the total similarity between $\vct{a}^i$ to all answers $\{\vct{a}^j\}_{j=1}^n$ becomes $\frac{P(C_k)}{Q} = \bar{P}(C_k)$ where $C_k$ is the semantic group of $\vct{a}^i$. Therefore, Eq~\ref{eq:walg} reads
    \begin{align}
        \alg(\vct{q}) &= - \sum_{i=1, \vct{a}^i \in C_k}^n \bar{P}(\vct{a}^i | \vct{q}) \log \bar{P}(C_k) \\
        &= - \sum_{i=1, \vct{a}^i \in C_k}^n \frac{\tilde{P}(\vct{a}^i | \vct{q})}{Q} \log \bar{P}(C_k) \\
        &= - \sum_{k=1}^M \frac{P(C_k)}{Q} \log \bar{P}(C_k) \\
        &= - \sum_{k=1}^M \bar{P}(C_k) \log \bar{P}(C_k) \\
        &= \se(\vct{q})
    \end{align}
\end{proof}

\section{Additional experimental details}\label{app:exp_details}
\textbf{Models.} We use Llama2-7B, Llama2-13B~\citep{touvron2023llama}, Llama-3.1-8B~\citep{dubey2024llama}, 
Phi-3-mini-4k-instruct~\citep{abdin2024phi}, gemma-2-2b-it~\citep{team2024gemma}, and Mistral-Nemo-Instruct-2407~\citep{jiang2023mistral}.

\textbf{Datasets.} We evaluate our method across three NLP tasks: question answering (QA), text summarization (TS), and machine translation (MT).
Following SE~\cite{farquhar2024detecting}, we cover four different QA categories: commonsense knowledge (\squad~\cite{rajpurkar2018know} and \triviaqa~\cite{joshi2017triviaqa}), general knowledge from Google search (\nq~\cite{kwiatkowski2019natural}), simple math problems (\svamp~\cite{patel2021nlp}), biology and medicine (\bioasq~\cite{krithara2023bioasq}).
For text summarization, we adopt the abstractive single-document summarization dataset (\xsum~\cite{narayan2018don}) and the email subject line generation dataset (\aeslc~\cite{zhang2019email}), as in LM-Polygraph~\citep{fadeeva-etal-2023-lm}.
For machine translation, we evaluate on two widely used datasets: WMT-14 German-to-English and WMT-14 French-to-English~\cite{bojar2014findings}.

\textbf{Baselines.} We compare our methods with white-box UQ methods including \klefull~\citep{nikitin2024kernel}, \se~\citep{farquhar2024detecting}, \naive~\cite{malinin2020uncertainty}, Eigenscore~\cite{chen2024inside}, SAR~\cite{duan2023shifting} and black-box UQ methods including \kleheat~\citep{nikitin2024kernel}, \dse~\citep{farquhar2024detecting}, \ptrue~\citep{kadavath2022language}, \lexsim~\citep{fomicheva2020unsupervised}, graph-based methods (\numset, \sumeigv, \degree, \ecc)~\citep{lin2024generating}, LUQ-Pair~\cite{zhang2024luq}.

\textbf{Generation setting.} 
For each question, we generate one answer at a low temperature $(T = 0.1)$ to assess model correctness and sample 10 answers at a high temperature $(T = 10)$ to estimate uncertainty. We use 5-shot in-context demonstrations for QA tasks and instruct the model to produce short answers following the brief prompt from \se. For TS and MT tasks, we employ a 0-shot setting and adopt the instructions from LM-Polygraph. Table~\ref{tab:instruct_prompt} provides a summary of the prompts used across different tasks.

\textbf{Correctness measures.} Following~\citep{farquhar2024detecting}, we evaluate QA tasks using the F1 score, a standard metric for the \squad~dataset, with a correctness threshold of 50\% applied across all QA datasets. For text summarization and machine translation, we use \rouge~\cite{lin-2004-rouge} and \bert~\cite{bert-score}, as in LM-Polygraph.

\textbf{Hyperparameter setting.} We select the best scale factor $\tau$ in Eq~\ref{eq:alg} and~\ref{eq:walg} from $\{ 0.1, 1, 10, 100 \}$. For similarity function $f$, we use \rouge~score.

\textbf{Implementation details.} For calculating \rouge, we use the Python implemention of Google~\href{https://github.com/google-research/google-research/tree/master/rouge}{https://github.com/google-research/google-research/tree/master/rouge}. For models and datasets, we download them from Hugging Face~\href{https://huggingface.co/}{https://huggingface.co/}.

\textbf{Computational resources.} We conduct each experiment three times using NVIDIA RTX A6000 GPUs.

\section{Additional experimental results}\label{app:exp_results}
\subsection{Detailed results}
\textbf{QA datasets.} Table~\ref{tab:qa_auroc_no_math} demonstrates the average AUROC score of 6 different models on 5 QA datasets. Across different model architectures, \alg~and \walg~are the best-performing white-box and black-box methods, respectively. \klefull~is the second-best white-box method but its black-box \kleheat~is surpassed by \dse~which is the runner-up in the black-box setting. In terms of the AUARC score, Table~\ref{tab:qa_auarc_no_math} shows that \alg~still yields the highest score in the black-box setting while \klefull~exhibits a competitive performance with \walg.

\textbf{TS and MT datasets.} Table~\ref{tab:ts_mt_prr} illustrates the average PRR score with respect to two different correctness measure: \rouge~and \bert. In the black-box scenario, \alg~shows the superior advantages over other baselines across different settings, followed by \lexsim. By capturing the semantic distance in the kernel space, KLE improves over (D)\se~in both white-box and black-box settings though they are still behind our methods by a clear margin. This reinforces the effectiveness of explicitly integrating the intra-and inter-cluster semantic similarities into entropy calculation.

\subsection{Ablation studies}
\textbf{Effect of similarity function on \alg.} We examine three different choices of similarity functions: \rouge, \entail, and \embed~in Section~\ref{sec:method}. For \entail~with NLI models, we use the same DeBERTa model~\cite{he2020deberta} that is used for semantic clustering. For \embed~with sentence transformer, we use the Qwen2-7B-instruct model~\cite{li2023towards} which ranked 1st in the MTEB benchmark~\cite{muennighoff2022mteb}. Table~\ref{tab:change_sim} summarizes the PRR score on TS and MT tasks. \rouge~results in the best performance except for translation task with \bert~in which \rouge~ties with \embed. The superior of \rouge~also explains the good performance of \lexsim~in Table~\ref{tab:ts_mt_prr}.

\textbf{Effect of scale factor $\tau$.} When $\tau$ is large, SNNE smooths out differences, making all intra-and inter-distances contribute more equally. In that case, SNNE behaves more like an average pairwise similarity similar to Deg or LexSim. In contrast, when $\tau \to 0$, it emphasizes the smallest ``intra-distance'' and ignores all other intra-and inter-distances. Additionally, in one-sentence output scenarios, and with a specific design of the similarity function $f$, our method asymptotically ($\tau \to 0$) recovers LUQ-Pair. Table~\ref{tab:scale_factor} illustrates the impact of varying scale factor on QA tasks in Figure~\ref{fig:main_qa_auroc} and TS \& MT in Figure~\ref{fig:main_ts_mt_prr} below. Our method is not sensitive to the choice of $\tau$ in the tuning set. Overall, $\tau = 1$ yields the best performance across different use cases, making it our default choice.

\end{document}